%% file: main.tex
\documentclass[letterpaper, 10 pt, conference]{ieeeconf}  

\usepackage{graphicx}
\usepackage{amsmath}
\usepackage{amssymb}
\usepackage{csquotes}
\usepackage{url}
\usepackage{hyperref}
\usepackage{multirow}
\usepackage{color, colortbl}
\usepackage{algorithm}
\usepackage{algpseudocode} 
\usepackage[nolist]{acronym} 
\let\labelindent\relax
\usepackage[inline]{enumitem} 
\usepackage{cite}
\graphicspath{ {imgs/} }
\usepackage{siunitx}

\newif\ifDB
\DBfalse

\newif\ifshowComments
\showCommentstrue

\ifshowComments
    \usepackage[draft,markup=bfit, deletedmarkup=sout, authormarkup=brackets]{changes}
\else
    \usepackage[final,markup=bfit, deletedmarkup=sout, authormarkup=brackets]{changes}
\fi

\definechangesauthor[name={Lukas},color=green]{LR}
\definechangesauthor[name={Matej}, color=orange]{MH}
\definechangesauthor[name={Karel}, color=blue]{KB}

\ifshowComments

\else

\fi

\input{acronyms.tex}

\input{macro.tex}

\useVspacefalse 

 \usepackage[firstpageonly=true]{draftwatermark}

\SetWatermarkAngle{0}
\SetWatermarkColor{black}
\SetWatermarkLightness{0.5}
\SetWatermarkFontSize{9pt}
\SetWatermarkVerCenter{30pt}
\SetWatermarkText{\parbox{30cm}{%
\centering This is the authors' final version of the manuscript published as:\\
\centering Bartunek, K., Rustler, L., \& Hoffmann, M. (2026),\\
\centering No Need to Look! Locating and Grasping Objects by a Robot Arm Covered with Sensitive Skin \\
\centering  in '2026 IEEE International Conference on Robotics \& Automation (ICRA)'. (C) IEEE \\
\centering
}}

\title{No Need to Look! Locating and Grasping Objects by a Robot Arm Covered with Sensitive Skin}

\ifDB

\author{Anonymous Authors%
}

\else

\author{Karel Bartunek$^*$, Lukas Rustler$^*$, and Matej Hoffmann %
\thanks{$^{*}$Both authors contributed equally.}%
\thanks{Karel Bartunek, Lukas Rustler and Matej Hoffmann are with the Department of Cybernetics, Faculty of Electrical Engineering, Czech Technical University in Prague,
 {\tt\small lukas.rustler@fel.cvut.cz, matej.hoffmann@fel.cvut.cz.}}
 \thanks{This work was co-funded by the European Union under the project Robotics and Advanced Industrial Production (reg. no. CZ.02.01.01/00/22\_008/0004590). L.R. was additionally supported by the Czech Science Foundation (GAČR) under project No. 26-22606S and Grant Agency of the Czech Technical University in Prague, grant No. SGS24/096/OHK3/2T/13.}}

\fi
 
\IEEEoverridecommandlockouts
\begin{document}

\maketitle

\begin{abstract}
Locating and grasping of objects by robots is typically performed using visual sensors. Haptic feedback from contacts with the environment is only secondary if present at all. In this work, we explored an extreme case of searching for and grasping objects in complete absence of visual input, relying on haptic feedback only. The main novelty lies in the use of contacts over the complete surface of a robot manipulator covered with sensitive skin. The search is divided into two phases: (1) coarse workspace exploration with the complete robot surface, followed by (2) precise localization using the end-effector equipped with a force/torque sensor. We systematically evaluated this method in simulation and on the real robot, demonstrating that diverse objects can be located, grasped, and put in a basket. The overall success rate on the real robot for one object was 85.7\% with failures mainly while grasping specific objects. The method using whole-body contacts is six times faster compared to a baseline that uses haptic feedback only on the end-effector. We also show locating and grasping multiple objects on the table. This method is not restricted to our specific setup and can be deployed on any platform with the ability of sensing contacts over the entire body surface. This work holds promise for diverse applications in areas with challenging visual perception (due to lighting, dust, smoke, occlusion) such as in agriculture when fruits or vegetables need to be located inside foliage and picked.    

\end{abstract}

\input{Sections/introduction}
\input{Sections/method}
\input{Sections/results}

\input{Sections/conclusion}

\bibliographystyle{IEEEtran}
\bibliography{refs}

\end{document}

%% file: acronyms.tex
\newacro{dnn}[DNN]{Deep Neural Network}
\newacro{fcn}[FCN]{Fully Convolutional Network}
\newacro{sdf}[SDF]{Signed Distance Function}
\newacro{cnn}[CNN]{Convolutional Neural Network}
\newacro{gnn}[GNN]{Graph Neural Network}
\newacro{dl}[DL]{Deep Learning}
\newacro{ml}[ML]{Machine Learning}
\newacro{gpis}[GPIS]{Gaussian Process Implicit Surface}
\newacro{mlp}[MLP]{Multi-Layer Perceptron}
\newacro{dof}[DoF]{Degree of Freedom}

\newacro{ai}[AI]{Artificial Intelligence}
\newacro{llm}[LLM]{Large Language Model}
\newacro{hri}[HRI]{Human-Robot Interaction}
\newacro{phri}[pHRI]{Physical Human-Robot Interaction}
\newacro{shri}[sHRI]{Social Human-Robot Interaction}

\newacro{slam}[SLAM]{Simultaneous Localization and Mapping}
\newacro{pf}[PF]{Particle Filter}
\newacro{kf}[KF]{Kalman Filter}

\newacro{ros}[ROS]{Robot Operating System}
\newacro{gui}[GUI]{Graphical User Interface}
\newacro{urdf}[URDF]{Unified Robot Description Format}

\newacro{ik}[IK]{Inverse Kinematics}

\newacro{rrmc}[RRMC]{Resolved-Rate Motion Control}
\newacro{yarp}[YARP]{Yet Another Robot Platform}

\newacro{eit}[EIT]{Electrical Impedance Tomography}
\newacro{gp}[GP]{Gaussian Process}

\newacro{ft}[F/T]{Force/Torque}

%% file: macro.tex
\newcommand{\equationref}[1]{\hyperref[#1]{Eq.~\ref*{#1}}}
\newcommand{\figref}[1]{\hyperref[#1]{Fig.~\ref*{#1}}}
\newcommand{\tabref}[1]{\hyperref[#1]{Table~\ref*{#1}}}
\newcommand{\secref}[1]{\hyperref[#1]{Section~\ref*{#1}}}
\newcommand{\algoref}[1]{\hyperref[#1]{Alg.~\ref*{#1}}}

\newcommand{\etal}[1]{#1 \textit{et al.}}

\newif\ifuseVspace

\def\airskin{AIRSKIN}

%% file: Sections/introduction.tex
\section{Introduction}

Perception for robot manipulation has been dominated by visual inputs from cameras (RGB) or depth cameras (RGB-D). Classical methods have been used for object segmentation and pose and shape estimation to feed the synthesis of grasp proposals for a robot hand (e.g., \cite{kragic2016_3Dvision}). Recently, deep learning approaches to grasping have become popular, with point clouds from depth cameras being the most popular input data format \cite{newbury2023deep}. Tactile sensors have played a marginal role, with occasional use of feedback from sensorized fingertips.

\begin{figure}[htb]
    \centering
    \includegraphics[width=0.75\columnwidth]{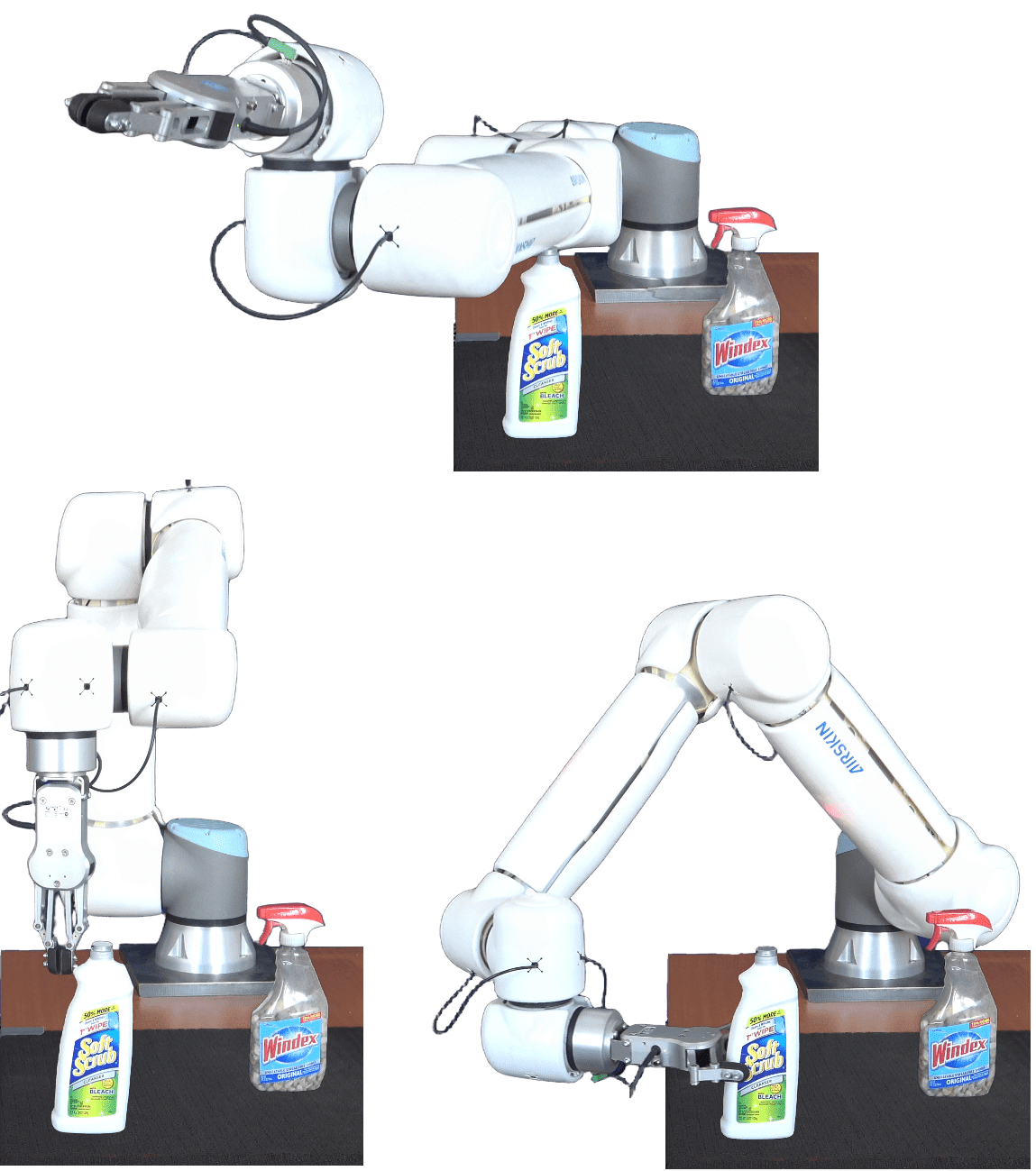}
    \caption{Rough scan with the whole-body skin (top), precise localization with end-effector (bottom-left) and pre-grasp position (bottom-right).}
    \label{fig:schema}
    \vspace{-1.75em}
\end{figure}

Humans approach perception for manipulation differently. They do not rely solely on vision and capitalize much more on the sense of touch---the active probing of the objects by hands with feedback from proprioception and touch. Touch also plays a role in locating the objects. We do not need precise object localization and often simply move our arms toward the object using peripheral vision until contact is established. 

This work examines an extreme case of this scenario with no visual feedback at all but exploiting the sense of touch over the entire surface of the arm. Imagine that you were to clean the table with your eyes closed. You would stretch your arm and move it around to quickly find the objects. After detecting them by touch, you would try to grasp the individual objects. 

The performance is intrinsically limited compared to the scenario with allowed vision, but it is an important capacity that we believe has been overlooked in robotics \cite{billard2025robots} and that can be applied in different application domains where vision is challenging, such as due to lighting, dust, or occlusion. 
The advent of large-area sensitive skins for robots (e.g., \cite{cheng2019comprehensive, maiolino2013FlexibleRobustLarge}), partly driven by their utility to enable or speed up safe human-robot collaboration (e.g., \cite{svarny2022effect}), makes it possible to deploy them as a new distributed touch sensor for robot manipulation. 

Robot haptic exploration has been studied mainly on robot hands with touch sensors on fingertips \cite{li2024ReviewOfTactileGrasping}. \etal{Jamali}~\cite{jamali2016ActivePerceptionBuildinga} used the fingertips of the humanoid robot iCub to actively explore the surfaces of objects using \ac{gp}~\cite{Williams2006} to model the surface and sampling points for tactile exploration on or near the estimated object boundary. \etal{Yi}~\cite{yi2016ActiveTactileObjecta} similarly used \ac{gp} to model the surface of individual objects explored with a sensor that outputs a single point of contact. In~\cite{sun2018CombiningContactForces} the authors combine haptic features such as friction and surface roughness with local geometry to recognize the object using the Gaussian-Bayesian classifier. 

Tactile sensors have also been used for workspace exploration and object localization. This task requires obtaining a large amount of global information about the workspace. \etal{Abraham}~\cite{abraham2017ErgodicExplorationUsing} combined binary sensing (collision / no collision) in combination with ergodic exploration\footnote{Ergodic trajectory optimization computes control laws to drive a dynamic system such that the amount of time spent in regions of the state space is proportional to the expected information gain in those regions~\cite{miller2016ErgodicExplorationDistributed}}. \etal{Kaboli} proposed a method~\cite{kaboli2017TactileBasedFrameworkActive} for locating and clustering objects in the workspace using a combination of tactile skin and proximity sensors. They then exploited this method to estimate physical properties, such as stiffness. In later work~\cite{kaboli2019TactilebasedActiveObjecta}, the authors showed a fully tactile approach that explores the workspace while reducing variance as quickly as possible (based on a model utilizing \acp{gp}) and used it to estimate the physical properties of objects in the workspace. The authors in~\cite{felip2012ContactbasedBlindGrasping} developed a simple yet effective system for exploring objects in a box using a three-finger gripper equipped with tactile sensors. Their proposed method searches for objects using linear movements and upon contact attempts to grasp them. More recently, the method proposed by \etal{Murali}~\cite{murali2020LearningGraspSeeingb} also performs linear scans but updates the pose of objects using \ac{pf}. They learned an auto-encoder network that is able to regrasp an object in the case of unsuccessful grasps. \etal{Pai}~\cite{pai2023TactoFindTactileOnlya} utilized a three-finger gripper to acquire tactile information about an object and search for the given object in clutter.

Whole-body sensitive skins, used in this work, have been used typically in the context of safe \ac{hri}---to detect and localize collisions and to keep impact forces within the prescribed limits by triggering the robot stop or other reactions ~\cite{svarny2022effect, luu2024TacLinkIntegratedRobotArm}. Jiang and Wong~\cite{jiang2024HierarchicalFrameworkRobot} further proposed a hierarchical framework to learn danger-free paths and also react appropriately upon contact using a real-world robot with artificial skin. In addition, when combining proximity and tactile whole-body sensing, robots can actively avoid obstacles and simultaneously reduce the force of impact in case of collision~\cite{caroleo2024ProxyTactileReactiveControl}.

Only a few works use touch to explore large parts of the robot workspace. \etal{Saund} proposed a theoretical formulation as the `Blindfolded Traveler's Problem', a variation on the famous planning problem~\cite{saund2022BlindfoldedRobotBayesian}. \etal{Albini}~\cite{albini2021ExploitingDistributedTactile} used large-scale skin (more than 3000 sensors) on the Baxter robot to reach a given goal when navigating through unknown obstacles. A similar task was performed by \etal{Meng}~\cite{meng2024ObstacleAvoidancePath}, when they utilized skin on one link of a robot to autonomously avoid obstacles while moving an object grasped by the gripper of the robot. 

The main contribution of this work is a practical demonstration that a `blindfolded robot' relying only on haptic feedback over its complete surface of the body can explore its workspace and find and grasp all the objects there. The method using whole-body contacts is six times faster compared to a baseline that uses haptic feedback only on the end-effector. We present a pipeline that uses a standard industrial robot UR10e, a parallel OnRobot RG6 gripper, and whole-body robotic skin \airskin{}~\footnote{\url{https://www.airskin.io/}} with coarse spatial resolution. This robotic setup is used to showcase the method. The method can be adapted to other setups with whole-body sensing and holds promise for a number of applications where visual perception for manipulation is challenging. The code is available at \url{https://rustlluk.github.io/nntl}.

%% file: Sections/method.tex
\section{Materials and Methods}
This section gives details on the hardware and software used and describes the pipeline and algorithmic solutions. The robot is mounted on a table (see \figref{fig:setup}). Before each experiment, selected objects from the YCB~\cite{Calli2015YCB} dataset are placed on top of the table.

\subsection{Robot, Skin and Simulation}
\begin{figure}[htb]
    \centering
    \includegraphics[width=0.49\columnwidth]{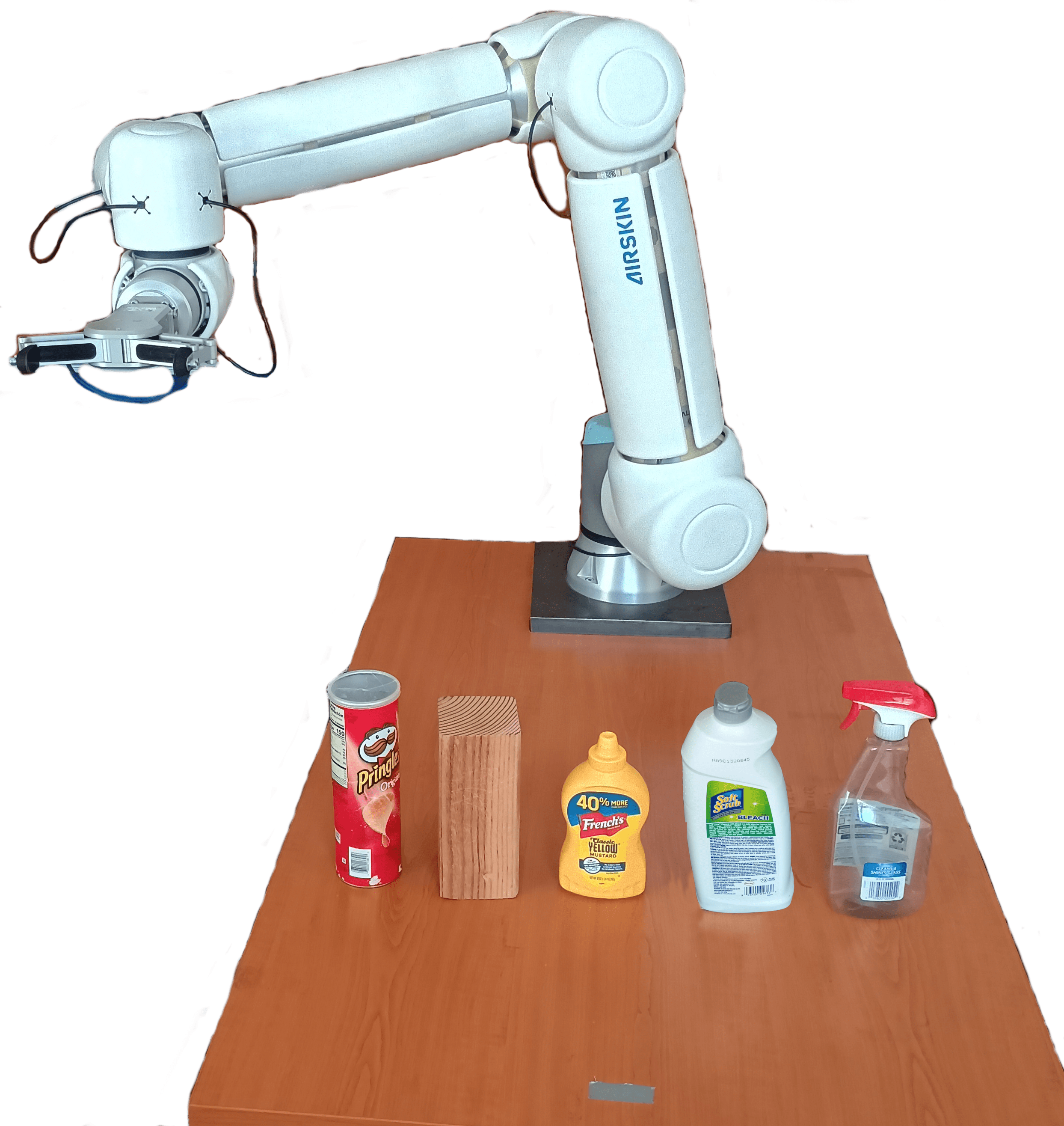}
    \includegraphics[width=0.49\columnwidth]{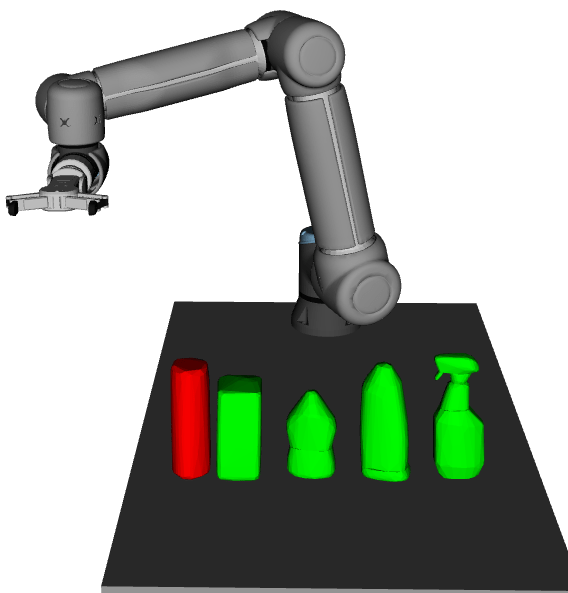}
    \caption{Robot with objects in the real world (left) and in simulation (right).}
    \label{fig:setup}
    \vspace{-0.5em}
\end{figure}

We tested our approach using an UR10e 6-axis robotic manipulator equipped with artificial skin \airskin{}, both in simulation and in the real world---see \figref{fig:setup}. The robot is a collaborative industrial manipulator that is used both in research and in industry. The skin is a commercially sold product that, in standard operation, is connected to the robot safety controller and sends a signal immediately after a change in pressure inside any of its parts---we call these pads---reaches a predefined threshold. In standard operation, this command stops the robot. We have the skin connected to a proprietary controller that allows us to read the pressures of each pad and process them without necessarily halting the robot. The spatial resolution of the skin is rather low---each skin pad (white skin parts in \figref{fig:setup}) is one sensor, and thus we know that the collision occurred on one of the 10 pads without any better localization. In addition, the custom controller reads the data with a frequency of \SI{30}{\hertz}. Both the robot and the skin are accessible and controlled through \ac{ros}.

In addition to the real robot, we also performed experiments in simulation. We utilized simulation created in~\cite{rustler2024adaptive}. The underlying simulation of physics is achieved by PyBullet~\cite{coumans2021} and simulation has a \ac{ros}-enabled interface, allowing us to implement the proposed method using the same code in both environments. 

\subsection{Blind Localization and Grasping}
\begin{figure}[htb]
    \centering
    \includegraphics[width=1\columnwidth]{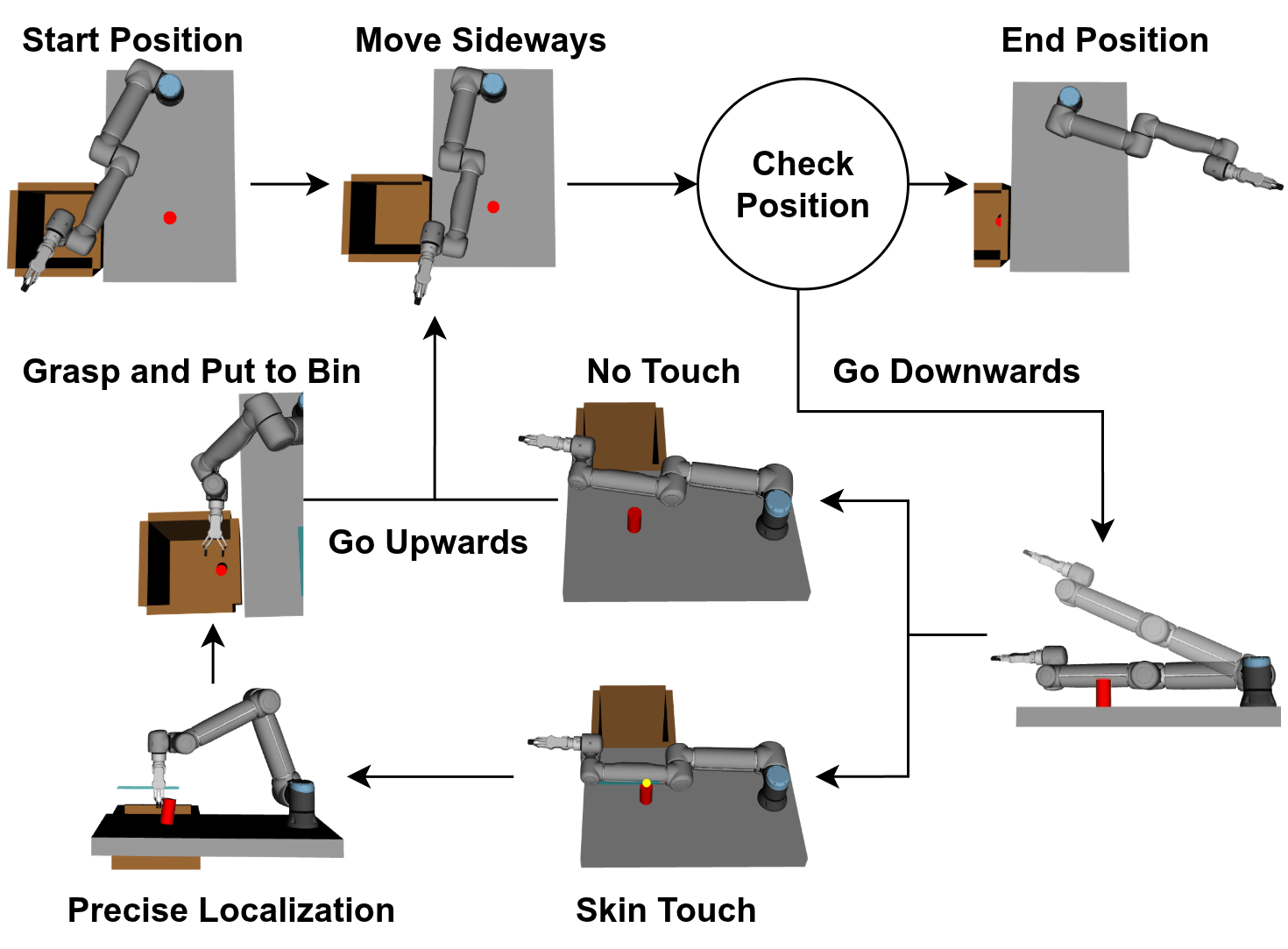}
    \caption{Schema of the exploration pipeline. The robot starts at the rightmost position with the arm fully stretched forward and starts moving sideways with a fixed step. If it is not at the end position, it moves downwards at each step. In case of a collision, precise localization begins, followed by a grasping operation. After the grasping ends, or the robot moves to the lowest position, it returns upwards. The loop continues until the robot reaches the end position.}
    \label{fig:schema}
\end{figure}
The pipeline is visualized in \figref{fig:schema}. The exploration starts with the robot being fully stretched in the start position---in our case on a right side of a table---raised above the objects. The robot then moves to the other side of the table with a fixed step. At every step, it descends towards the table. The state of the artificial skin is continually examined during this movement and, if a collision is detected, the robot is stopped, and the precise localization (see \secref{sec:localization}) and the grasping module (see \secref{sec:grasping}) are run. After the grasping module finishes, or after the robot reaches the table, it returns to the position before the downward movement and continues to the next step sideways. Execution stops when the robot reaches the leftmost position. See also the accompanying video at \url{https://rustlluk.github.io/nntl}, where several runs of the pipeline are shown.

We decided to go from right to left. However, this decision is arbitrary and the system would function the same using the opposite direction. We assume that 
\begin{enumerate*}[label=(\roman*)]
    \item all movements (sideways movements, precise localization movements, and grasp approaches) are performed from one side to explore objects one by one;
    \item all objects can be grasped using a lateral grasp;
    \item there is enough space for the robot to maneuver;
    \item the objects are tall enough to be detected by the robot---in our case, we apply a safety measure that stops the robot 10\,cm above the table.
\end{enumerate*}

\subsection{Object Localization}
\begin{figure}[htb]
    \centering
    \begin{minipage}[t]{1\columnwidth}
        \centering
        \includegraphics[width=0.8\textwidth]{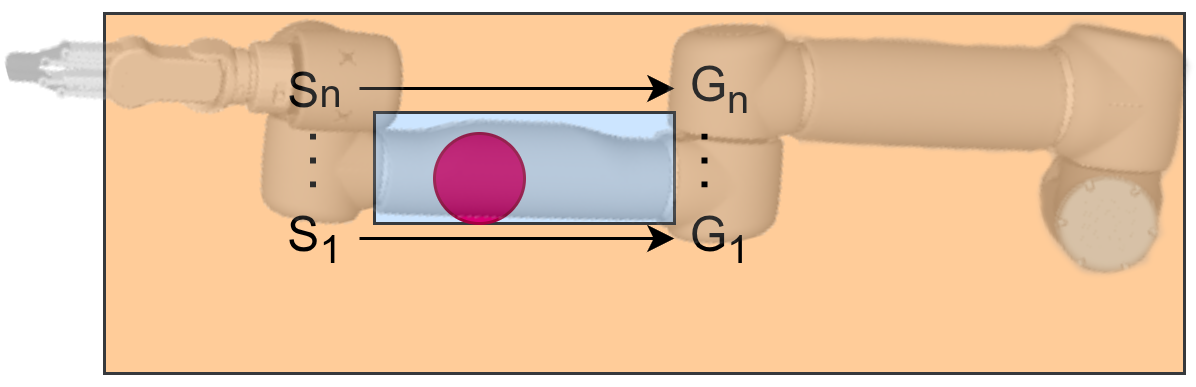}
        
        (a)
        \label{fig:top_views_scan}
    \end{minipage}
    
    \vspace{0.5cm} 
    \begin{minipage}[t]{0.49\columnwidth}
        \centering
        \includegraphics[width=0.7\textwidth]{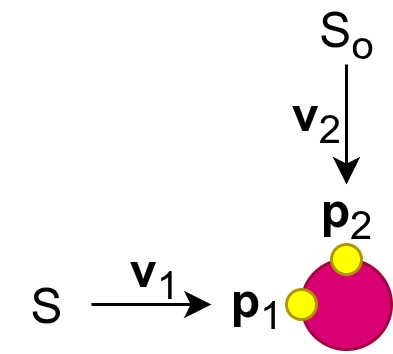}
        
        (b)
        \label{fig:top_views_orth}
    \end{minipage}\hfill
    \begin{minipage}[t]{0.49\columnwidth}
        \centering
        \includegraphics[width=0.6\columnwidth]{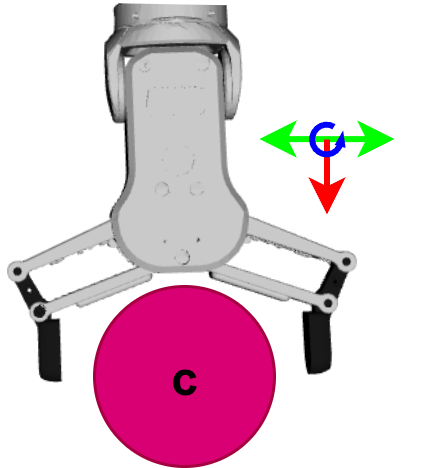}
        
        (c)
        \label{fig:top_views_grasping}
    \end{minipage}
    \caption{(a) Top-view on table (orange), object (red) and projected area of the pad in contact (blue). $S_n, G_n$ are n-th start and end positions of the precise localization. (b) Top-view of the object (red) with both contacts $\mathbf{p}_1, \mathbf{p}_2$ (yellow) with approach vectors from both initial ($S, \mathbf{v}_1$) and orthogonal ($S_o, \mathbf{v}_2$) direction. (c) Top-view on the initial grasp pose. The colored coordinate frame shows directions for grasp poses refinement.}
    \label{fig:top_views}
    \vspace{-1.5em}
\end{figure}

\label{sec:localization}
We propose a localization method divided into two parts: 
\begin{enumerate*}[label=(\roman*)]
    \item rough scan,
    \item and precise localization.
\end{enumerate*}
The rough scan utilizes the artificial skin. When contact of a given skin pad with an object is detected, a 2D rectangle created by the dimensions of the given pad is projected onto the table based on the current pose of the robot (the robot approaches from above)---see \figref{fig:top_views_scan}. This rectangle represents the upper bound of an area where the object is located. However, the individual skin pads are quite large---the projection for the smallest pad is a square with a side of 12\;cm, for the largest it is a rectangle with sides of 40 and 15\;cm. These dimensions are then also slightly inflated to account for objects that hit the edge of a single pad.

\begin{figure}[htb]
\vspace{-1em}
\begin{algorithm}[H]
\caption{Precise Localization}
\label{alg:precise_loc}
    \begin{algorithmic}[1] 
    \Statex \textbf{Input:} 2D rectangular projection area $\mathbf{P}$ of pad in contact; width of the gripper $w$
    \Statex \textbf{Output:} 2D contact points $\mathbf{p}_1, \mathbf{p}_2$; 2D direction vectors $\mathbf{v}_1, \mathbf{v}_2$ 
    
    \State $S, G \gets $ \textit{GetStartEnd($\mathbf{P}, w$)};
    \While{\textit{InsideArea($\mathbf{P}, S, w$)}}
        \State $\mathbf{p}_1, \mathbf{v}_1 \gets $ \textit{ScanWithContactDetection($S, G$)};
        \label{line:scan_1}
        \If{$\mathbf{p}_1$ is not None} \Comment{If contact occurred}
            \State $S_o, G_o \gets$ \textit{GetOrthStartEnd($\mathbf{P}, w, S, G, \mathbf{p}_1$)};
            \State $\mathbf{p}_2, \mathbf{v}_2 \gets $ \textit{ScanWithContactDetection($S_o, G_o$)};
            \If{$\mathbf{p}_2$ is not None}
                \State \textbf{Return: }$\mathbf{p}_1, \mathbf{p}_2, \mathbf{v}_1, \mathbf{v}_2$
            \EndIf
        \Else
            \State $S, G \gets$ \textit{GetStartEnd($\mathbf{P}, w, S, G$)};
    \EndIf
    \EndWhile
    \State \textbf{Return: } $\emptyset$
    \end{algorithmic}
\end{algorithm}
\vspace{-1.5em}
\end{figure}

The precise localization is achieved with Cartesian-linear movements of the end-effector in the rectangular area of the pad in contact (see \figref{fig:top_views_scan})---the pseudocode is in \algoref{alg:precise_loc}.
The robot moves from the left utmost rear position (S$_1$ in \figref{fig:top_views_scan}) towards the base of the robot (G$_1$ in \figref{fig:top_views_scan}). The contact probe (end-effector) points to the table in the height of the lowest point of the pad in contact. The \acf{ft} sensor in the wrist of the robot is used to detect collisions between the robot and the object. If no collision is detected, this is repeated for S$_2$,G$_2$ and so forth. The distance between S$_n$ and S$_{n+1}$ is the effective width of the gripper side that faces the objects during this movement. 
The localization finishes if a collision is detected during movement or if the rightmost scan in the rectangle is completed (movement from S$_n$ to G$_n$ in \figref{fig:top_views_scan}). If a contact is detected, the end-effector pose during the collision is recorded as $\mathbf{p}_1$ along with the current direction of movement $\mathbf{v}_1$ and the robot performs a corrective Cartesian-linear movement orthogonal to the previous one---movement from S$_o$ in \figref{fig:top_views_orth}. If contact is detected during this movement, the contact point $\mathbf{p}_2$ and the direction of movement $\mathbf{v}_2$ are recorded. If there is no contact, $\mathbf{p}_1$ is discarded, and the process continues with the same starting and ending position. See the accompanying video at \url{https://rustlluk.github.io/nntl}
for a visual representation.

\subsection{Grasping}
\label{sec:grasping}

After finding the contact points and approach vectors, the approximate center of the object can be found as an intersection of two rays, i.e., $\mathbf{c} = \mathbf{r}_1 \cap \mathbf{r}_2$. Here we define a ray as a line
\begin{equation}
    \mathbf{r}_n = \mathbf{p}_n + s\mathbf{v}_n,
\end{equation}
where $s \in \mathbb{R}$ is a parameter, $n \in \bigl\{1,2\bigr\}$, and $\mathbf{p}_n, \mathbf{v}_n$ are the points and vectors obtained during precise localization. The center point is 2D (in the table plane). The third coordinate needed for grasp (the grasp height) is set to be the minimum height of the skin pad in contact.

The initial grasp pose $G$ is composed of the center $c$ and an orientation. For the initial grasp, its orientation is chosen so that the gripper jaws are parallel to the table and the distance between each finger and the object center is the same---see \figref{fig:top_views_grasping}. The initial approach vector is identical to $\mathbf{v}_2$.

The center point is an estimation of the object's center of mass with uncertainty coming from the lack of information about the object's shape (only two points of the surface are used), low resolution of the skin pads and noise in both the skin and the \ac{ft} sensor. Therefore, if the initial grasp fails, we test several grasp poses $\mathbf{G}_i$, each being a fixed 2D pose transformation of the initial grasp projected onto the table, consisting of the following parameters (from the point-of-view of the gripper): 
\begin{enumerate*}[label=(\roman*)]
    \item distance behind the center (0 to 40\;mm; red in \figref{fig:top_views_grasping});
    \item distance to the sides (-45 to 45 \;mm; green in \figref{fig:top_views_grasping});
    \item rotation around the center (-30 to 30 degrees; blue in \figref{fig:top_views_grasping}).
\end{enumerate*}
The poses are tested sequentially until a successful grasp is detected---and then the object is moved to a bin---or when the pose $\mathbf{G}_{i_{max}}$ fails to yield a successful grasp. 

The real gripper has an inbuilt method that detects failed grasps. We combine this with the current width of the gripper, i.e., when the current width is lower than a given threshold, we know that there cannot be any object. In the case of the simulated gripper, we combine the current width of the gripper with information about whether the fingers are in contact with an object.

\subsection{Baseline Method}
\label{sec:baseline_localization}
To justify the benefits of whole-body skin, we employed a simple baseline method. The method is based on the assumption that there is no whole-body sensing available and that the only contact sensor is the \ac{ft} sensor in the wrist. We scan the whole workspace using only Cartesian-linear movements. The method is the same as the precise localization in \ref{sec:localization}, but now the explored area is the entire table.

%% file: Sections/results.tex
\section{Experiments and Results}
The task in all the experiments was the same: Find all objects on a table, grasp them, and put them away. We performed multiple experiments with various objects in distinct poses in simulation and in the real world. 

We selected five objects from the YCB dataset~\cite{Calli2015YCB}. The objects can be seen in \figref{fig:setup}. We selected the objects to represent a diverse sample of shapes. In the real world, we put a rubber mat on top of the table to increase friction. In addition, we filled the hollow objects with small stones to increase their weight, making them heavy enough to be detected by the \ac{ft} sensor. The weights of the objects range circa from 450\;g to 1000\;g, based on their size and shape.

\subsection{Effect of Whole-Body Skin}
To evaluate the effect of whole-body skin, we compared our method with a simple baseline that explores the entire workspace only with the end-effector---see \secref{sec:baseline_localization}. We performed five runs of ours and the baseline method and compared the runtimes---in all cases, the object was successfully grasped and cleaned from the table. The results can be seen in \tabref{tab:baseline_comparison}. We can see that the effect of the skin is enormous. In the case without object (rough scan), our method is about 13 times faster (87\,s versus 1135\,s). With one object on the table, our method is 6 times faster (197\,s versus 1198\,s). The decrease in performance with one object is caused by the fact that our method needs to perform the complete precise localization and grasping when contact is detected. On the other hand, the baseline only needs to perform the second part of precise localization (contact from the orthogonal direction) and grasping. However, using whole-body skin significantly decreases the time required.

\begin{table}[htb]
\caption{Comparison of runtimes for exploration with end-effector only (baseline) and with whole-body skin (ours). The values are computed over 5 runs.}
\label{tab:baseline_comparison}
\centering

\resizebox{\columnwidth}{!}{%
\begin{tabular}{c|c|c|c|c|}
\cline{2-5}
 & \textbf{\begin{tabular}[c]{@{}c@{}}Baseline\\ No Object\end{tabular}} & \textbf{\begin{tabular}[c]{@{}c@{}}Baseline\\ Object\end{tabular}} & \textbf{\begin{tabular}[c]{@{}c@{}}With Skin\\ No Object\end{tabular}} & \textbf{\begin{tabular}[c]{@{}c@{}}With Skin\\ Object\end{tabular}} \\ \hline
\multicolumn{1}{|c|}{\textbf{Time {[}s{]}}} & 1135 & 1198 & 87 & 197 \\ \hline
\end{tabular}%
}

\end{table}

Note that during the baseline experiments we needed to decrease the sensitivity of the \ac{ft} sensor to avoid false detection during long Cartesian-linear movements along the table.



\subsection{Effect of Object Position in Workspace}
\label{sec:obj_position}
\begin{figure}[htb]
    \centering
    \includegraphics[width=1\columnwidth]{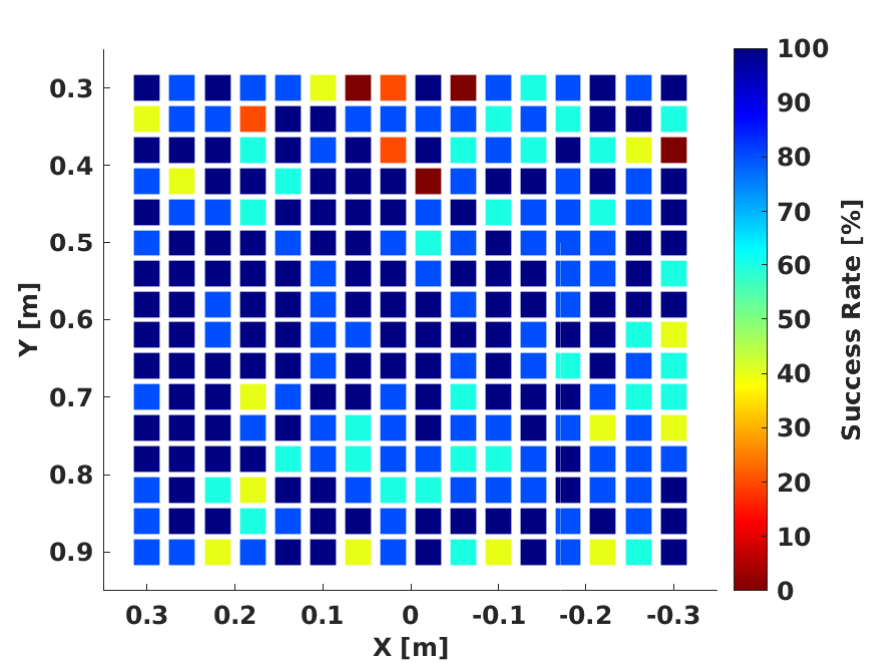}
    \caption{Success rate for cylindrical object (red object in \figref{fig:setup}) in simulation at different positions in the workspace. Each square represents one position of the cylinder as its center. The values are computed over 5 repetitions at every location. The base of the robot is located at position (0, 0).}
    \label{fig:position_variability}
\end{figure}

The first experiment of our method aims to show how the performance of the pipeline changes based on the position of the object in the workspace. We selected a cylindrical can (the red object in \figref{fig:setup}), as it is symmetric around the Z-axis, and thus is orientation invariant. We placed the object in a grid of $60\times60$\;cm (a limitation of the table and the maximum range of the robot) with a step of 4\;cm, resulting in 256 positions and 1280 repetitions in total (5 repetitions for each position). Repetitions were considered completed after the algorithm explored the entire workspace or after 10 minutes. As this experiment is time-consuming, it was performed only in simulation. The results are shown in \figref{fig:position_variability}.

We can see that around the middle of the workspace, the success rate is in most cases at least 80\% (4 out of 5), with few entries of 60\% (3 out of 5). At the positions closest to the robot base---the base is at (0, 0) in the workspace---the success rate drops and occasionally reaches 0\%. Other places with a low success rate are around the boundaries of the workspace. However, in both cases, this is expected as the manipulability in these areas is much lower, and certain configurations are not reachable by the robot in desired orientations. Simulation glitches and inaccuracies might also contribute, especially since we have not manually checked all individual runs. Nevertheless, the average success rate of this experiment in simulation of 83.1\% indicates that the method demonstrates robust position invariability.

\subsection{Effect of Object Shape and Orientation}
\label{sec:obj_shape}
The second experiment aims to show the impact of the shape of the objects on the results. We used all the objects shown in \figref{fig:setup} and tested them one by one. The experiment was performed in both simulation and the real world. The majority of the objects are not symmetric, and thus we tested them in five different orientations---90, 45, 0, \mbox{-45}, and \mbox{-90} degree rotations around their Z-axis. The cylindrical object used in previous experiment is symmetric along the Z-axis and thus only one rotation was considered. The default orientation (0 degrees) for each object can be seen in \figref{fig:setup}. All objects were placed in a position with coordinates (0\,m, 0.6\,m) in \figref{fig:position_variability}---in the middle of the table's shorter dimension and 0.6\,m away from the base in the longer dimension. In total, 21 experiments were performed with 10 repetitions in both environments, resulting in 210 runs for each environment.

The average success rate in simulation is 75.7\%. It is lower than in the previous experiment (83.1\%), but the task is now arguably more difficult, with the shape of the object being more diverse. The results suggest that the pipeline can be generalized to different shapes.

\begin{figure}[htb]
    \includegraphics[width=0.95\columnwidth]{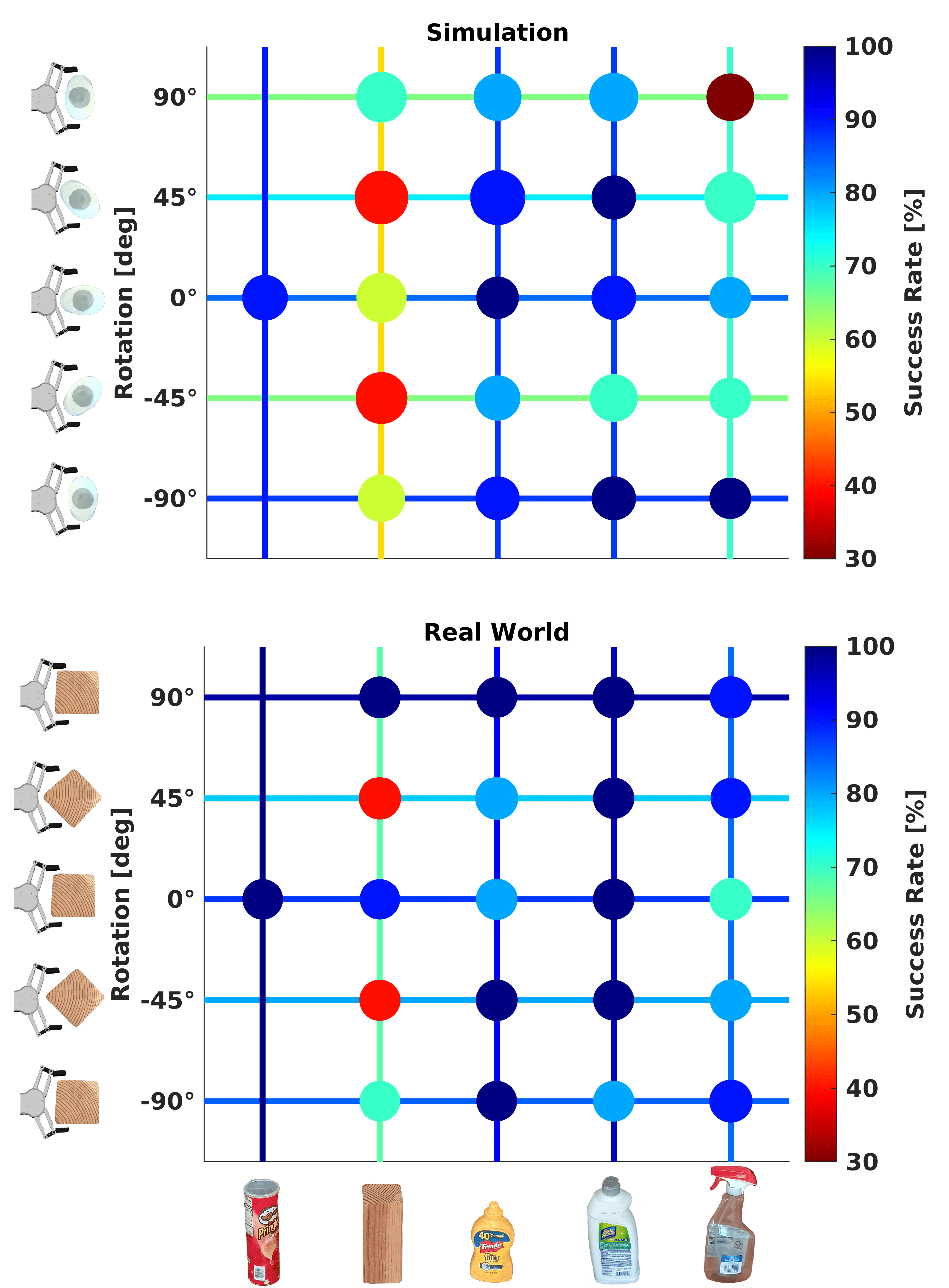}
    \caption{Success rate of grasping objects with different rotations in simulation (top) and real world (bottom). Color of the circles corresponds to the success rate and the their size to the average duration---in simulation the smallest circle is about 220\,s and the biggest one about 390\,s and in the real world 210\,s and 235\,s, respectively. The times are computed over successful runs. The color of the vertical lines show mean success rate per object and vertical lines per rotation. The objects in rotation of 0 degrees are depicted under the plot and two objects in all rotations are shown on the left side. The first object is symmetric along the Z-axis and thus only one rotation was considered. The values are computed over 10 repetitions.}
    \label{fig:center}
    \vspace{-1.5em}
\end{figure}

The results from simulation are shown in the top plot of \figref{fig:center} and the most noticeable feature is that the success for the second object (block) is the lowest (slightly over 50\% in average). It is a counterintuitive result, as the block is a simple shape from a human perspective. We believe that the dimensions of the block are the major contributing factor to the block being difficult to grasp. Our gripper has a maximum width of 15\,cm and the block has a side of 8.5\,cm. As the estimated center of the localized object is usually not precise, the object may not always fit between the gripper jaws. We observed that 
\begin{enumerate*}[label=(\roman*)]
    \item the gripper fingers often hit the block, and thus moved it instead of getting it between the jaws;
    \item the block got between the jaws, but the grip was not strong enough and the object fell out of the gripper during movement.
\end{enumerate*}
 In the plot, we can see that the situation is even worse for the $\pm45^\circ$ rotations, as the practical width of the block is its diagonal---12\,cm. The effect of width is visible when looking at the results for $0^\circ$ rotation. Grasping in this rotation is intuitively the most reliable, as the narrowest side of the object faces in the direction of the first grasp attempt. The lowest success rate among the rest of the objects was 80\% for this rotation. This shows that the overall shape influences the performance only slightly if the objects are rotated conveniently. 
 
If we inspect the other rotations, we can see that there is a clear difference. The results differ between objects, but there is a clear trend showing that $0^\circ$ and $\pm 90^\circ$ achieve the best results. This result is a natural consequence of how our pipeline works. As we only use two perpendicular points to determine the center of the object, we cannot infer any direct information about the shape or orientation of the object. Thus, in the case of $\pm 45^\circ$ rotations, the initial grasp is not aligned with any principal axis of the object. To grasp successfully, objects must either slide on the table to align with the gripper, be deformable enough to be squeezed, or one of the alternative grasp poses must relatively align. Using more points of contact is required for proper shape and pose estimation. The results are also supported by the time needed to successfully clear the object---the size of circles in \figref{fig:center}. We can see that for $\pm 45^\circ$ the system takes more time on average than for the other, with the rotation of $0^\circ$ being the fastest on average. The average runtime over all objects for in rotation of $0^\circ$ is 262\,s, while it is 311\,s for the $\pm 45^\circ$ rotations.

In the real world, the average success rate is 85.7\%, which is 10\% higher than in simulation. We attribute it to the deformability of the real-world objects (except the wooden block), whereas the deformability was not implemented in simulation. Therefore, the gripper can create a better grip than in simulation by squishing the object inside the fingers.

When we focus on individual objects, the trends are close to those in simulation. The worst performance is achieved for the block, followed by the Windex bottle. For the block, the $0, 90, -90^\circ$ rotations achieved a higher success than in simulation. We believe that it is caused by better friction between the block and the fingers of the gripper and by better grasp failure detection by the real gripper. The performance of the $\pm45^\circ$ rotation is similar, as it is not influenced by only its width, but also because grasps of cubes in $\pm45^\circ$ angles from its faces tend to be less stable. Typically, the object was grasped in two neighboring faces, causing the object to slip off the gripper. The runtimes in the real world are generally lower and more similar between objects than in simulation, ranging from 205\,s to 235\,s. The cause is the main discrepancy between the environments, which is the behavior of the object after unsuccessful grasp. In the real world, objects usually fall over, and thus become ungraspable. In simulation, the objects more often just move along the table. This can be seen from the number of grasp attempts. The system attempted to grasp $2.2\pm1.9$ and $1.2\pm0.6$ times in simulation and in the real world, respectively.

Overall, the real-world system can remove one object from the table with a success rate of more than 85\% in less than 4 minutes (218\,s) on average.

\subsection{Effect of Clutter}

\begin{figure}[htb]
    \centering
    \includegraphics[width=0.95\columnwidth]{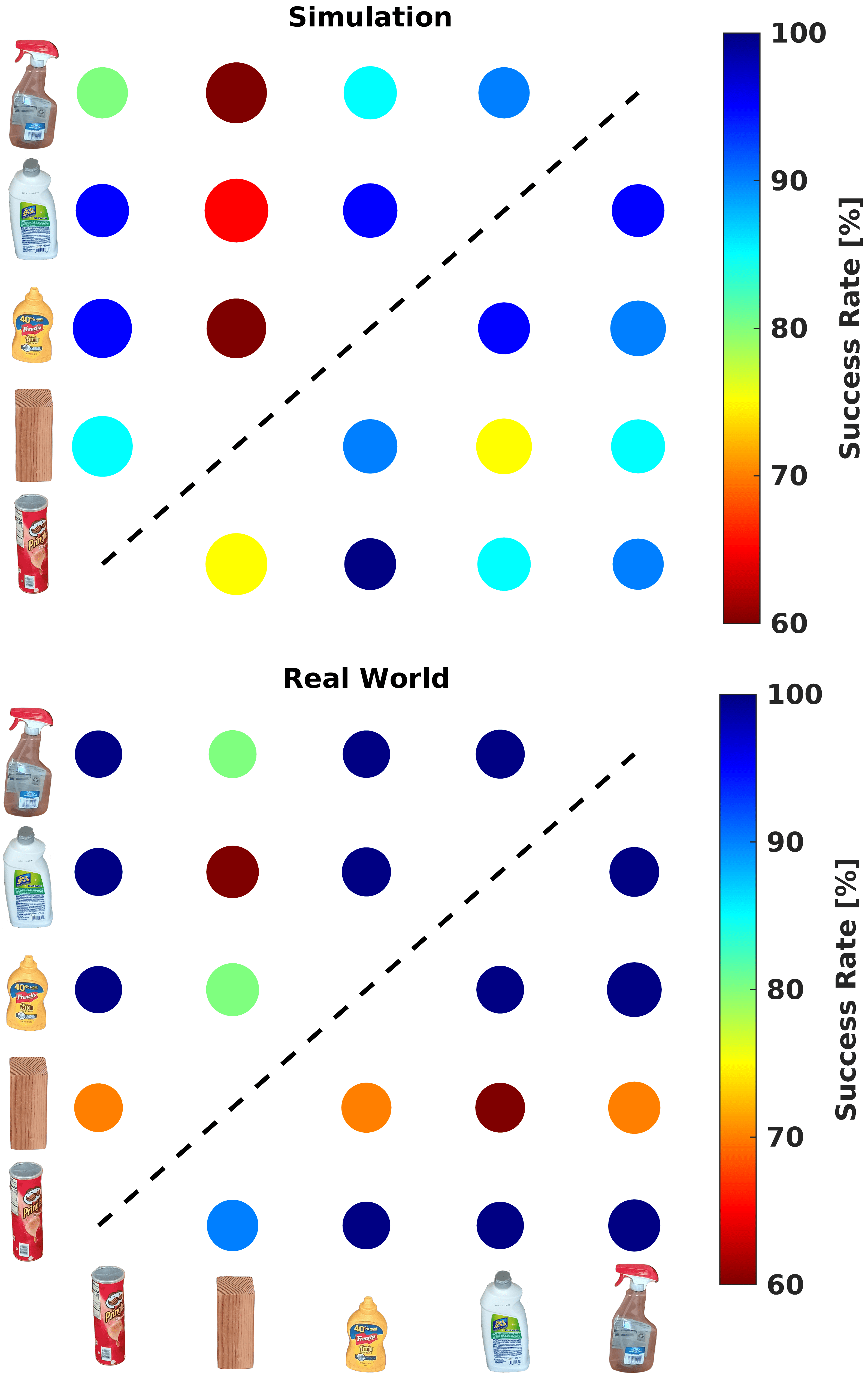}
    \caption{Success rate of experiment with two object on a table at a time in simulation (top) and real world (bottom). Color corresponds to the success rate and the size of circles to the average duration---in simulation the smallest circle is about 340\,s and the biggest one about 525\,s and in the real world 290\,s and 390\,s, respectively. The times are computed over runs where at least one object was successfully removed from the table. For each column-row pair the object in column was in contact with the robot first. The values are computed over 10 repetitions in simulation and 5 in the real world.}
    \label{fig:two}
\end{figure}
\label{sec:multiple}

This experiment evaluates the ability to handle more than a single object in the scene. This experiment should answer the questions of whether our system is able to handle a more cluttered space and whether it is robust enough to complete the rest of the task even when one of the objects falls over or fails otherwise.

We tested scenarios with two and three objects in the scene. In simulation, we tested all possible permutations of two and three objects, resulting in 80 different settings in total---20 for two objects and 60 for three objects. In the real world, we tested all 20 permutations for the case of two objects. Due to time constraints, we tested only 10 combinations--all triplets with random order of objects---in the case of three objects at a time.

We have chosen two reliable positions for the objects represented by coordinates (X, Y) as (0.25\,m, 0.75\,m) and (-0.2\,m, 0.55\,m) with respect to \figref{fig:position_variability}. All objects are placed with a rotation of $0^\circ$ to avoid orientation-based variance. We acknowledge that using the best rotation (based on the previous section) influences the results. However, this experiment focuses on how the pipeline handles multiple objects while reducing other sources of uncertainty, and thus we believe that this is a fair comparison. 

The results of the two-at-a-time experiment are in \figref{fig:two}. The success rate is defined as the total number of objects in the basket at the end and not whether all objects are in the basket, i.e., when one of two objects is in the basket, the success rate is 50\%.

In the case of simulation (top), the average success rate is 85.5\%, showing that the pipeline is generally able to handle more than one object. If we look more closely, we can see that overall there is no difference in object order except for the block. If the block is placed such that it is in contact with the robot first (second column of the plot), the average success rates range from 60 to 75\%. In the other case (second row), the performance is 75 to 90\%. This is caused by the fact that if the block is second, there is a greater change in the placement of at least one object successfully. We can also check the average duration, which is the highest for the second column ($420\pm280$\,s compared to experiment average of $365\pm175$\,s). That, again, corresponds to more grasp attempts. The average number of grasps (over both objects in the scene) for the second columns is $3.3\pm2.3$, while for the whole experiment it is $2.5\pm1.3$.

In the real world, the average success rate is 89\%. It is again higher than in simulation, similar to the previous experiment. The main difference from simulation results is that the performance of individual pairs is now 100\% more often.

\subsubsection{Three objects}
\begin{figure}[htb]
    \centering
    \begin{minipage}{\columnwidth}
        \centering
        \includegraphics[width=1\textwidth, trim={0cm 0cm 0cm 0cm},clip]{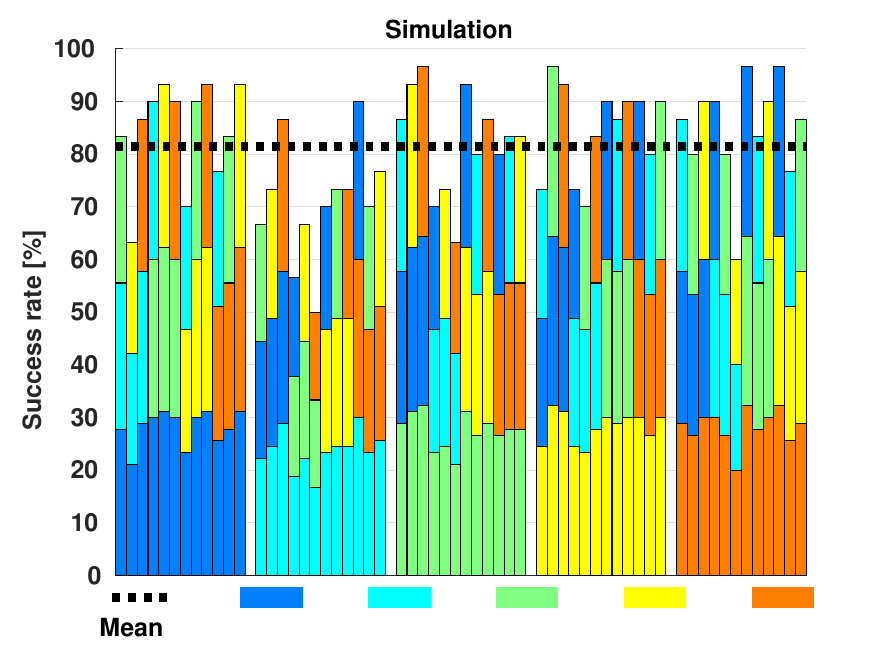}
        
        \vspace{-0.9em}
        \hspace{4.25em}
        \includegraphics[height=1.15cm]{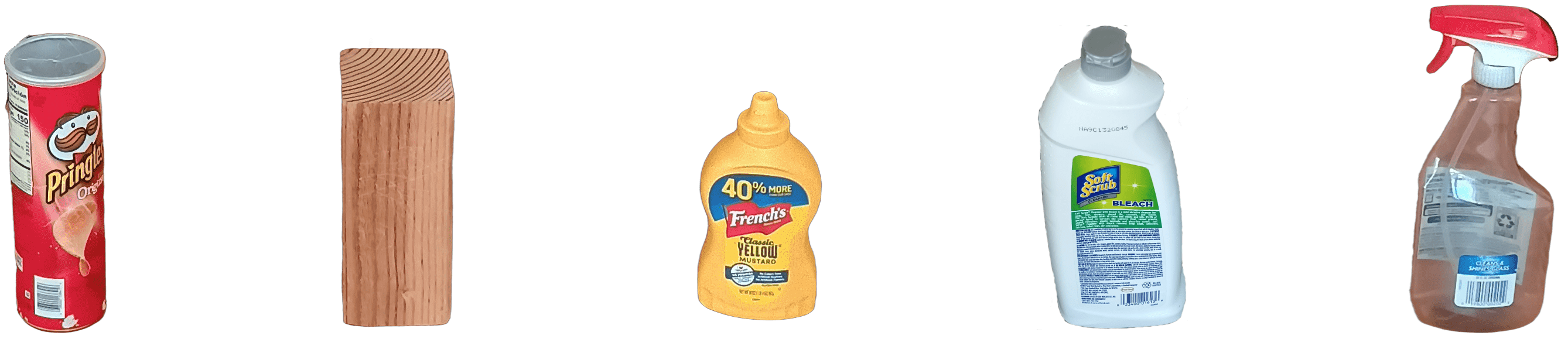}
    \end{minipage}
    \caption{Success rate of experiment with three object on a table at a time in simulation. The colors of bar segments serve to show which objects were used based on the legend. The order is from the bottom, i.e., the bottom object was the first in contact with the robot. The values are computed over 10 repetitions.}
    \label{fig:three}
    \vspace{-1.5em}
\end{figure}

Lastly, performance was evaluated with three objects present at the table at the same time. The objects were placed in the three positions used in previous experiments. The average success rate of this experiment is 81.5\% in simulation and 88.0\% in the real world. In simulation, the total success rate is less than 5\% lower than in the case of two objects at a time. In the real world, it is just 1\% lower. The detailed results from simulation can be seen in \figref{fig:three}---we do not show real results as we only tested a small subset of combinations of objects. The plot does not show any effect of the order of objects. Except for the case where block is the first, where the average success rate is the lowest.

%% file: Sections/conclusion.tex
\section{Conclusion and Discussion}

We presented a system for locating and grasping objects in the complete absence of vision, exploiting tactile feedback over the complete surface of a robot arm only. The full results are available in \tabref{tab:final}. Using a simple cylindrical object in simulation, we showed that the system is able to locate and grasp objects anywhere in the workspace with a success rate of 83\%. Furthermore, in both simulation and the real world we have shown that our proposed framework can handle objects with different shapes and different poses with respect to the robot and the workspace. 
Our experiments demonstrated that the system is able to successfully remove different objects in more than 75\% cases in simulation and 85\% cases in the real world---in under four minutes on average. The experiments further demonstrate that the pipeline is more effective for deformable objects. Experiments with 2 and 3 objects in the workspace at the same time suggest that our pipeline can handle even cluttered environments, achieving more than 80\% in simulation and almost 90\% in the real world. We compared our method with a simple baseline method that does not use the whole-body skin. The results showed that using whole-body skin to scan the workspace is about six times faster than using only the end-effector of the robot. 
Recordings of the experiments and the code are publicly available at \url{https://rustlluk.github.io/nntl}.

\begin{table}[htb]
\centering
\caption{Results of different experiments---position variability~\secref{sec:obj_position}, shape variability~\secref{sec:obj_shape}, and multiple objects at a time~\secref{sec:multiple}. The first experiment was done only in simulation.}
\label{tab:final}
\resizebox{\columnwidth}{!}{%
\begin{tabular}{c|c|c|c|c|}
\cline{2-5}
 & \textbf{\begin{tabular}[c]{@{}c@{}}Position\\ Variability\end{tabular}} & \textbf{\begin{tabular}[c]{@{}c@{}}Shape\\ Variablity\end{tabular}} & \textbf{\begin{tabular}[c]{@{}c@{}}Two objects\\ at a time\end{tabular}} & \textbf{\begin{tabular}[c]{@{}c@{}}Three Objects\\ at a time\end{tabular}} \\ \hline
\multicolumn{1}{|c|}{\textbf{Simulation}} & 83.1\% & 75.7\% & 85.5\% & 81.5\% \\ \hline
\multicolumn{1}{|c|}{\textbf{Real World}} & - & 85.7\% & 89.0\% & 88.0\% \\ \hline
\end{tabular}%
}

\end{table}

Our results show the feasibility and benefits of whole-body robot workspace exploration with only haptic feedback. 
Although the objects used in the experiment have generic shapes, they still share some key characteristics that allow the manipulator to execute precise localization actions. Furthermore, all evaluated objects are tall and heavy enough to be detected by the artificial skin.

The whole-body skin used had a very coarse spatial resolution on the body. Moreover, the mounting of the skin did not allow for horizontal exploratory sweeps in the workspace, and hence we resorted to vertical movements. Higher spatial resolution would increase the accuracy of object localization. We also observed hardware failures resulting in temporal false negatives for contact detection, one of which is shown in the accompanying video. For precise localization, we used a \ac{ft} sensor on the robot wrist. This resulted from the physical limitations of our setup. However, this sensor can be omitted if a different sensitive sensor is available, such as tactile sensors on the hand.

In general, we consider the reduced generality of our pipeline acceptable. Although it is a significant limitation of our approach, it is a direct consequence of the physical constraints of the platform. Future work will focus on formalizing general approaches for tactile-only manipulation, along with the development of more advanced localization algorithms